%% file: m1854.tex
\documentclass{ecai} 

\usepackage{latexsym}
\usepackage{amssymb}
\usepackage{amsmath}
\usepackage{amsthm}
\usepackage{booktabs}
\usepackage{enumitem}
\usepackage{graphicx}
\usepackage{color}
\usepackage[normalem]{ulem}
\usepackage{algorithm2e}
\usepackage{subfig}
\usepackage{multirow}

\newcommand{\indep}{\perp \!\!\! \perp}
\newcommand{\dep}{\not\!\perp\!\!\!\perp}

\newtheorem{theorem}{Theorem}
\newtheorem{lemma}[theorem]{Lemma}

\newtheorem{proposition}[theorem]{Proposition}

\newtheorem{definition}{Definition}

\newcommand{\BibTeX}{B\kern-.05em{\sc i\kern-.025em b}\kern-.08em\TeX}

\begin{document}


\begin{frontmatter}


\paperid{1854} 


\title{Enabling Causal Discovery in Post-Nonlinear Models with Normalizing Flows}


\author{\fnms{Nu}~\snm{Hoang}\thanks{Corresponding Author. Email: nu.hoang@deakin.edu.au}}
\author{\fnms{Bao}~\snm{Duong}}
\author{\fnms{Thin}~\snm{Nguyen}} 

\address{Applied Artificial Intelligence Institute (A\textsuperscript{2}I\textsuperscript{2}), Deakin University, Australia.}


\begin{abstract}
Post-nonlinear (PNL) causal models stand out as a versatile and adaptable framework for modeling intricate causal relationships. However, accurately capturing the invertibility constraint required in PNL models remains challenging in existing studies. To address this problem, we introduce CAF-PoNo (Causal discovery via Normalizing Flows for Post-Nonlinear models), harnessing the power of the normalizing flows architecture to enforce the crucial invertibility constraint in PNL models. Through normalizing flows, our method precisely reconstructs the hidden noise, which plays a vital role in cause-effect identification through statistical independence testing. Furthermore, the proposed approach exhibits remarkable extensibility, as it can be seamlessly expanded to facilitate multivariate causal discovery via causal order identification, empowering us to efficiently unravel complex causal relationships. Extensive experimental evaluations on both simulated and real datasets consistently demonstrate that the proposed method outperforms several state-of-the-art approaches in both bivariate and multivariate causal discovery tasks.
\end{abstract}

\end{frontmatter}


\input{intro} 
\input{related} 
\input{preliminary} 
\input{methodology}

\input{experiments}

\input{conclusion}


\bibliography{references}

\onecolumn
\renewcommand\thesubsection{\Alph{subsection}}

\input{appendix}

\end{document}

%% file: intro.tex
\section{Introduction}

The need to uncover causal relationships from solely observational data where randomized controlled trials are impractical has led to the emergence of causal discovery methodologies as a crucial field of study. 
A key challenge in causal discovery stems from the non-uniqueness
of causal models that can induce the same data distribution  \cite{spirtes2001causation}. In other words,
recovering the causal structure becomes impossible without making
additional assumptions about the causal model. To overcome this, various
functional causal models (FCMs) have been proposed with ensured identifiabilities,
such as the linear non-Gaussian acyclic model (LiNGAM) \cite{shimizu2006alinear},
additive noise model (ANM) \cite{hoyer2008nonlinear}, and post-nonlinear
(PNL) model \cite{zhang2009onthe}. While LiNGAM is limited to 
linear relationships, ANM adheres to the assumption of the additive
noise with non-linear causal mechanisms.

Among them, the PNL model stands out for its generality
in modeling complex non-linear causal systems. Specifically, under PNL models, the effect $Y$ is generated from its cause $X$ via the structural equation $Y := g\left(h(X) + \epsilon_Y\right)$, where $\epsilon_Y$ is an exogenous noise independent of the cause. Here, $h(\cdot)$ can be any function but $g(\cdot)$ is constrained to be \textit{invertible}, which has consequently introduced critical challenges in modeling and estimation. Several approaches have been investigated to effectively estimate PNL models \cite{zhang2009onthe, uemura2020estimation, keropyan2023rankbased}, yet representing the exact invertibility required by the model still poses a major difficulty. For example, AbPNL \cite{uemura2020estimation} adopts auto-encoders to model a function and its ``pseudo inverse'' by minimizing the reconstruction error, so the full invertibility can only be obtained with zero reconstruction error everywhere, which is virtually impossible to achieve. Meanwhile, in \cite{keropyan2023rankbased} ranked-based methods are explored
as a means to estimate strictly monotonic (and thus invertible) functions, however their analyses focus more on linear inner functions.


\begin{figure}[t]
\includegraphics[width=1\columnwidth]{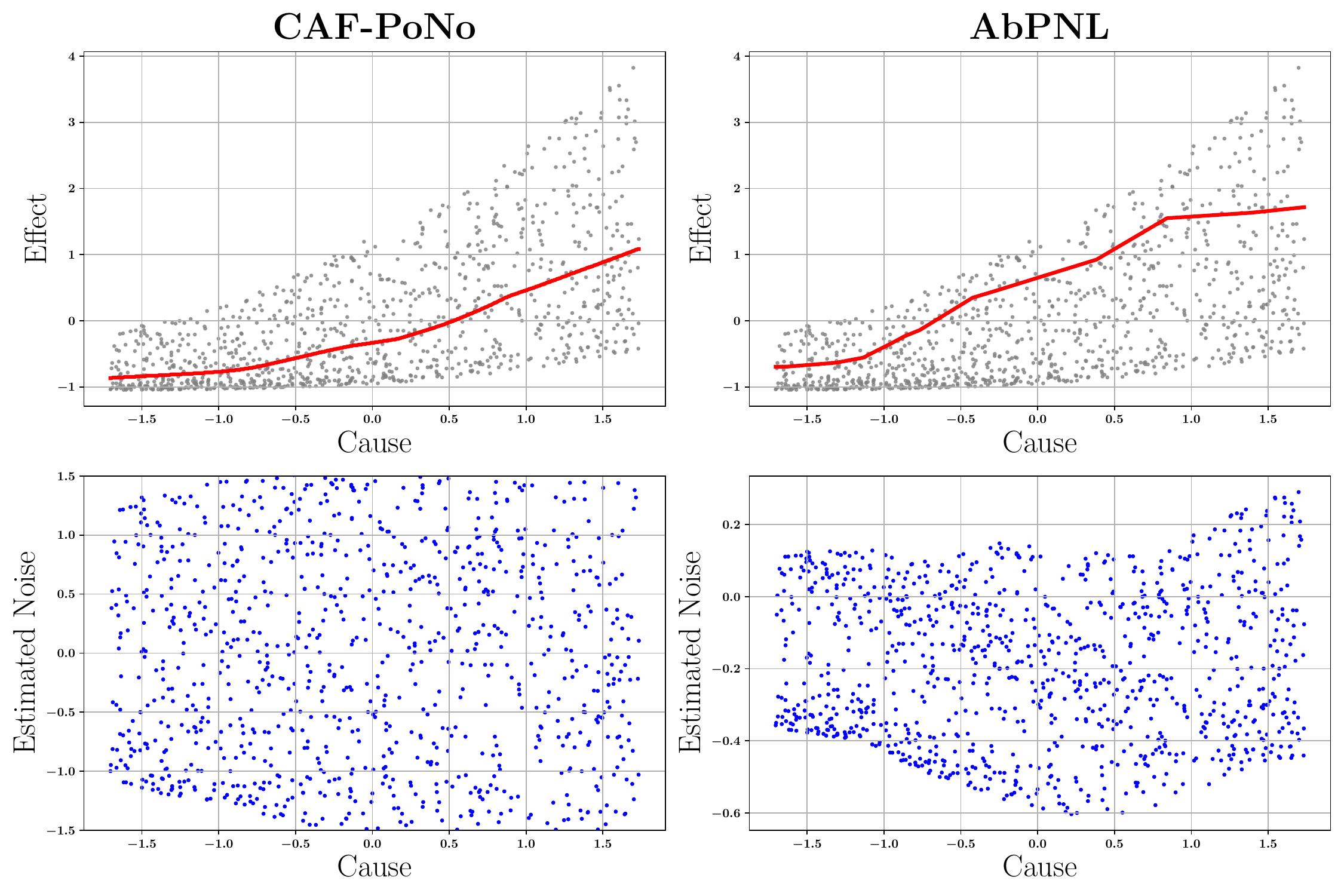}

\caption{Comparison between CAF-PoNo and AbPNL \cite{uemura2020estimation}
in cause-effect inference under PNL data. The red lines depict the estimated nonlinear functions based on the observed values of the effect and cause variables. In this example, AbPNL recovers the noises incorrectly due to the inability to model invertiable functions of auto-encoders, leading to the noise visibly dependent on the cause. Meanwhile, with the adoption of normalizing flows, our CAF-PoNo method can accurately capture the invertibility constraint and recover the noises more correctly, which are independent of the cause.
\newline
\vspace{2mm}
}
\label{fig:teasure}
\end{figure}

In this paper, to overcome the aforementioned challenges, we propose CAF-PoNo (\textbf{\uline{Ca}}usal discovery via Normalizing
\textbf{\uline{F}}lows for \textbf{\uline{Po}}st-\textbf{\uline{No}}nlinear
models)\footnote{Source code is available at \url{https://github.com/htn274/CAFPoNo}.}, a novel estimation method for causal discovery under PNL models that
capitalizes on the benefits of Normalizing Flows (NF), a powerful neural architecture known for its ability to model invertible functions \cite{papamakarios2021normalizing}. This innovative adoption of NFs allows for parametrizing highly complex PNL models with the invertibility of $g(\cdot)$ guaranteed without any ad-hoc enforcement, and thus drastically improves statistical efficiency in estimating them.
The NF-enabled PNL models are then used to recover the underlying noises for the subsequent cause-effect identification task via statistical independence measures, in which our CAF-PoNo method exhibits remarkable capabilities compared with existing methods.
As an example, Figure~\ref{fig:teasure} demonstrates CAF-PoNo's ability to correctly retrieve the underlying noise that is independent of the cause, whereas the autoencoder-based approach in AbPNL \cite{uemura2020estimation} suffers from the inaccurate invertible function estimation, resulting in an unfaithful noise that is still dependent on the cause.
Furthermore, the proposed approach excels in addressing both bivariate
and multivariate causal discovery, showcasing its effortless extensibility.
Our main contributions are summarized as follows:
\begin{enumerate}
\item We address the long-standing challenge of accurately modeling the invertible functional mechanism in PNL causal models with the introduction of the CAF-PoNo method. By leveraging normalizing flows, our method can capture the exact invertibility of the PNL model, leading to improved statistical efficiency and estimability compared with existing PNL-based methods. To the best of our knowledge, this is the first application
of normalizing flows to learn causal relationships under the PNL model.
\item We extend the CAF-PoNo framework from handling bivariate causal discovery (i.e., telling apart cause and effect) to handling multivariate causal discovery, by leveraging the causal ordering technique aided with independence scoring, exhibiting an attractive polynomial runtime with provable correctness.
\item We demonstrate the efficacy of the proposed CAF-PoNo method in causal discovery via an extensive set of numerical evaluations on both simulated and real
datasets.
The empirical results confirm the superiority of CAF-PoNo over competitive
baselines in both bivariate and multivariate scenarios.
\end{enumerate}
\textbf{}

%% file: related.tex

\section{Related works}


\subsection{Causal Discovery under PNL Models}

While various identifiable Structural Equation Models (SEMs) and estimation methods have been proposed
to address the identifiability issue in causal discovery \cite{shimizu2006alinear, hoyer2008nonlinear, zhang2009onthe, khemakhem2021causalautoregressive, immer2023onthe}, our study focuses on the PNL model, which is one of the most flexible model representing complicated causal relationships. Nevertheless, causal discovery under PNL models is very challenging and under-studied due to the invertibility constraint.

Several approaches have been proposed to address the challenge, yet they own certain limitations that hinder their ability to achieve higher performance. For instance, in \cite{zhang2009onthe}, the PNL model is modeled by neural networks and estimated by minimizing the
mutual information between the cause and noise.
However, this approach ignores the invertibility
constraint during the optimization process, which potentially results in invalid estimates that violate the invertibility requirement. To address this problem, AbPNL \cite{uemura2020estimation} leverages auto-encoders to represent ``approximately invertible'' functions, which are estimated by minimizing both the reconstruction loss and mutual information of the noise and the presumed cause. However, AbPNL suffers from two key limitations. Firstly, it lacks a guarantee of learning a truly invertible function, leading to failures in reconstructing the underlying noise in the effect variable, inevitably causing incorrect causal direction identification. Secondly, the loss function necessitates computing mutual information during training, resulting in significant computational costs, especially when dealing with multiple variables. Meanwhile, instead of jointly learning both $g(\cdot)$ and $h(\cdot)$, Keropyan
et al. \cite{keropyan2023rankbased} employs ranked-based regression methods
to separately learn the two transformations. Nevertheless, the study mainly focuses on linear settings for the inner function.

\subsection{Multivariate Causal Discovery}

Traditional causal discovery methods can be divided into three groups:
constraint-based methods \cite{spirtes2001causation,richardson1996modelsof,colombo2012learning},
score-based methods \cite{cussens2011bayesian,cussens2017bayesian,hemmecke2012characteristic},
and hybrid methods \cite{tsamardinos2006themaxmin,ogarrio2016ahybrid}.
Most of these methods search for the causal directed acyclic graph
(DAG) within the combinatorial space of graph structure, which is known to be challenging due to the
super-exponential explosion in complexity with respect to the number of variables \cite{robinson1977counting}.
This has changed in 2018 when Zheng et al. \cite{zheng2018dagswith} introduced a novel
approach known as NOTEARS that maps the discrete space to a continuous
one, enabling the utilization of various gradient-based optimization
techniques. This breakthrough opened up new possibilities for causal
discovery by incorporating deep learning and continuous optimization
methodologies \cite{yu2019daggnndag,lorch2022amortized,lorch2021dibsdifferentiable}. 

Alternatively, instead of searching the vast space of all possible causal structures,
another line of research focuses on finding a causal topological
order \cite{rolland2022scorematching,sanchez2023diffusion,yang2023reinforcement},
accomplished by searching over the space of permutations, which is
orders of magnitude smaller than the space of DAGs. By fixing a topological
order, the acyclicity constraint is naturally ensured, eliminating
the need for additional checks. After obtaining a causal topological
order, an additional pruning step is necessary to remove spurious
edges. This pruning process helps refine the inferred causal relationships
by discarding unnecessary connections between variables.

\subsection{Normalizing Flows in Causal Discovery}

Normalizing flows \cite{papamakarios2021normalizing}
have emerged as a class of expressive generative models within deep
learning, capable of effectively modeling complex probability distributions
by learning invertible transformations. In the context of causal discovery,
many studies have applied normalizing flows to encapsulate intricate data
distributions that may arise in practice\cite{ren2021causaldiscovery,cundy2021bcdnets,lippe2022efficient}.
For instance, a recent study \cite{khemakhem2021causalautoregressive} have employed autogressive flows in the context of location-scale (LS) models. More specifically, the causal mechanism for LS models is defined as $y=g\left(x\right)\epsilon+h(x)$, which deviates significantly from PNL models. Based on the formulations of PNL and LS models, they cannot generalize over each other, meaning that the techniques used in \cite{khemakhem2021causalautoregressive} are not applicable to PNL models, as empirically confirmed in Section~\ref{sec:Experiments}. To the best of our knowledge, normalizing flows have yet been studied for causal discovery under PNL models, where the invertible causal mechanism is of central importance, which highlights the significance of our study.


%% file: preliminary.tex
\section{Preliminary}
Suppose we observe an empirical dataset of $n$ i.i.d. samples of the
random vector $X=[x_{1},x_{2},...,x_{d}]^{\top}$ of $d$ variables. We assume that the data is generated under the PNL model as follows
\begin{equation}
x_{i}:=g_{i}\left(h_{i}\left(X_{\mathrm{Pa}(i)}\right)+\epsilon_{i}\right),i=1,2,...,d\label{eq:PNL_multi-1}
\end{equation}
 where $\mathrm{Pa}(i)$ is a set of direct causes (or parents) of
$x_{i}$. The noise variables $\{\epsilon_{i}\}$ are mutually independent
and therefore $\mbox{\ensuremath{\epsilon_{i}\indep X_{\mathrm{Pa}(i)}}}$
for each $\mbox{\ensuremath{i=1,2,...,d}}$. The corresponding causal
graph, denoted as $\mathcal{G}$, has directed edges $(j\rightarrow i):j\in\mathrm{Pa}(i)$
representing the causal relationships between variables. Similar to previous studies \cite{keropyan2023rankbased,lorch2022amortized,rolland2022scorematching},
this work also relies on assumptions of acyclicity and sufficiency.
That means the causal graph $\mathcal{G}$ is a directed acyclic graph (DAG) and there is no
hidden confounder. The goal of causal discovery is to
infer the true DAG $\mathcal{G}$ based on the empirical data of
$X$. 

For bivariate causal discovery, with a slight abuse of notation, we denote the cause and effect as $X$ and $Y$, respectively, where: 
\begin{equation}
y:=g(h(x)+\epsilon_{Y})\label{eq:PNL}
\end{equation}
The PNL model is identifiable
in most cases, except for specific instances where certain combinations
of functions and noise distributions are involved as outlined in \cite{zhang2009onthe}. Moreover, the PNL model is one of the most generic SEMs as
it generalizes both the prevalent LiNGAM \cite{shimizu2006alinear} and ANM \cite{hoyer2008nonlinear} models. 



%% file: methodology.tex
\section{CAF-PoNo: Causal Discovery via Normalizing Flows for Post-Nonlinear Models}

In this section, we first elaborate how to incorporate normalizing flows to design an effective estimator of PNL causal models. Then, we explain how to leverage this estimator for distinguishing cause and effect. Lastly, the framework is generalized to unravel the causal structure among multiple variables under PNL data.

\subsection{Normalizing Flows for the PNL model}

Normalizing flows is a powerful tool to express complex probability
distributions from a simple base distribution through several invertible
and differentiable transformations by exploiting the change of variables
rule. The primary objective of normalizing flows research is to devise
efficient transformations that possess the desirable properties of
invertibility and differentiability with a tractable derivative/Jacobian for effectively applying the change of variables rule.
For more details regarding normalizing flows, we refer to \cite{papamakarios2021normalizing}. 

Here, consider an empirical dataset of two variables $(X,Y)$ generated by Eq.~\eqref{eq:PNL}, we are interested in estimating the functions $g(\cdot)$ and $h(\cdot)$, so that we can recover the noise as

\begin{equation}
\epsilon_Y=g^{-1}(y) - h(x).
\label{eq:noise}
\end{equation}

Normalizing flows are used in this study specifically to model the invertible function $g(\cdot)$. In particular, we employ the cumulative distribution function (CDF)
flow \cite{papamakarios2021normalizing}, which is a simple, yet efficient invertible
transformation for one dimensional (scalar) variables. A CDF flow
is a positively weighted combination of a set of CDFs of any arbitrarily
positive density function. For example, the following mixture is a CDF, and is thus invertible:
\begin{equation}
f(x)=\sum_{i=1}^{k}w_{i}\Phi_{i}(x,\mu_{i},\sigma_{i}),\label{eq:CDF}
\end{equation}
where $k$ is the number of components, $w_{i}\geq0$, $\sum_{i=1}^{k}w_{i}=1$ and $(\mu_{i},\sigma_{i})$
are the weights and parameters for each component, respectively, whereas $\Phi_{i}$
is the cumulative distribution function of the $i$-th component. For simplicity, in our implementation, we consider a Gaussian distribution with mean $\mu_i$ and variance $\sigma_i^2$ for each component, yet it should be noted that \textit{our method is not restricted to Gaussian data} but can be applied to more generic settings. Thanks to the universal
approximation capability of Gaussian mixture models, the Gaussian
mixture CDF flow can express any strictly monotonic $\mathbb{R}\rightarrow(0,1)$
map with arbitrary precision \cite{ian2016deeplearning}. 

To apply the CDF flow for PNL estimation, let $\mbox{\ensuremath{z=h(x)+\epsilon_{Y}}}$,
then we have $\mbox{\ensuremath{y=g(z)}}$ and $\mbox{\ensuremath{z=g^{-1}(y)}}$.
We use neural networks to parametrize the inner function $h$, while CDF flows are used to model $\mbox{\ensuremath{g^{-1}}}$ instead
of $\mbox{\ensuremath{g}}$ since the noise directly relates to $\mbox{\ensuremath{g^{-1}}}$ as shown in Eq.~\eqref{eq:noise}. To model $\mbox{\ensuremath{g^{-1}}}$, which
is a $\mathbb{R}\rightarrow\text{\ensuremath{\mathbb{R}}}$ map, we
start with a CDF flow, which has a $(0,1)$ co-domain, then wrap it with another invertible map $(0,1)\rightarrow\mathbb{R}$,
such as an inverse sigmoid function. In short, our proposed NF-based approximation of $g^{-1}$ can be represented as follows:

\begin{equation}
z=g^{-1}(y)=\text{\ensuremath{\mathrm{sigmoid}}}^{-1}\left(\sum_{i=1}^{k}w_{i}\text{\ensuremath{\Phi_{i}}}\left(y,\mu_{i},\sigma_{i}\right)\right)\label{eq:sigmoid_CDF}
\end{equation}
and the noise can be estimated as: $\hat{\epsilon}_{Y}=z-h_{\theta}(x)$. 


\RestyleAlgo{ruled}
\begin{algorithm}[t]
\caption{CAF-PoNo algorithm.\label{alg:bivariate-causalinference}}

\KwIn{Empirical samples $(X, Y) = \{(x_i, y_i)\}_{i=1}^n$.}
\KwOut{The predicted causal direction.}
\SetKwFunction{CAF}{CAF-PoNo}

Standardize X and Y to zero mean and unit variance. \; 
Calculate the independence scores of two possible causal directions:  
$$ S_{X\rightarrow Y}=\CAF{$X$,$Y$}  $$ 
$$ S_{Y\rightarrow X}=\CAF{$Y$,$X$} $$\\

Determine the causal direction based on the independence scores:  
$$ \mathrm{dir}=\begin{cases} 			1, & \text{if $S_{X\rightarrow Y}>S_{Y\rightarrow X}$}\\             -1, & \text{if $S_{X\rightarrow Y} < S_{Y\rightarrow X}$} 		 \end{cases} $$

\KwRet{$\mathrm{dir}$} 

\BlankLine

\SetKwFunction{FMain}{CAF-PoNo}     
\SetKwProg{Fn}{Function}{:}{}     
\Fn{\FMain{$X$, $Y$}}{         
\KwData{Empirical samples for the cause $X = \{x_i\}_{i=1}^n$ and the effect $Y = \{y_i\}_{i=1}^n$.} 
\KwResult{The independence score for the causal direction $ X \rightarrow Y $.}
Perform train-test split to get $(X_\mathrm{train}, Y_\mathrm{train})$ and $(X_\mathrm{test}, Y_\mathrm{test})$.\;                  Train the CDF Flow on $(X_\mathrm{train}, Y_\mathrm{train})$ to obtain the learned parameters $\theta$.\;
        Estimate the noise:         $$          \hat{\epsilon}_{Y_{\mathrm{test}}}=g_{\theta}^{-1}(Y_{\mathrm{test}})-h_{\theta}(X_{\mathrm{test}})         $$\\
        Compute the independence score:         $$         S_{X\rightarrow Y}:=-\mathrm{HSIC}(\hat{\epsilon}_{Y_{\mathrm{test}}},X_{\mathrm{test}})         $$\\
        \KwRet{$S_{X\rightarrow Y}$}     }     \textbf{End Function}
\end{algorithm}

We adopt the Maximum Likelihood Estimation (MLE) framework to learn
the proposed PNL estimation in Eq.~\eqref{eq:sigmoid_CDF}. Let $\theta$
be the total set of all parameters, which includes the parameters
for the neural networks $h_{\theta}(x)$ and those for the CDF flow.
The CDF flow consists of a fixed number of $k$ Gaussian components,
with each component having its own set of parameters $w,\mu,\sigma$.
We chose Gaussian distribution as a base distribution for $\epsilon_{Y}\sim\mathcal{N}(0,1)$.
By the change of variables rule, the likelihood $p_{\theta}(y\mid x)$
can be expressed as follows:

\begin{align}
p_{\theta}(y\mid x) & =p_{\theta}(\epsilon_{Y}\mid x)\left|\frac{\partial\epsilon_{Y}}{\partial y}\right|\\
 & =p_{\epsilon_Y}(\epsilon_{Y})\left|\frac{\partial g_{\theta}^{-1}}{\partial y}\right|\\
 & =p_{\epsilon_Y}(g_{\theta}^{-1}(y)-h_{\theta}(x))\left|\frac{\partial g_{\theta}^{-1}}{\partial y}\right|
\end{align}

We aim to maximize the log-likelihood of the observed $Y$ conditioned
on $X$ over the space of $\theta$:
\begin{align}
\mathcal{L}(\theta) & =\frac{1}{n}\sum_{i=1}^{n}\ln p_{\theta}(y_{i}\mid x_{i})\nonumber \\
 & =\frac{1}{n}\sum_{i=1}^{n}\left(\ln p_{\epsilon_{Y}}(g_{\theta}^{-1}(y)-h_{\theta}(x))+\ln\left|\frac{\partial g_{\theta}^{-1}}{\partial y}\right|\right)\label{eq:mle_loss}
\end{align}
 where $\{(x_{i},y_{i})\}_{i=1}^{n}$ is the set of $n$ observed
samples of $(X,Y)$.

An important advantage of the employed CDF flow is the ability to tractably calculate the log-derivative $\ln\left|\frac{\partial g^{-1}}{\partial y}\right|$ in a closed-form fashion. Particularly, let $\mbox{\ensuremath{t=\sum_{i=1}^{k}w_{i}\text{\ensuremath{\Phi}}\left(y,\mu_{i},\sigma_{i}\right)}}$, we then have $z=-\ln(\frac{1}{t}-1)$, which yields: 

\begin{align}
\ln\left|\frac{\partial g^{-1}}{\partial y}\right| & =\ln\left|\frac{\partial g^{-1}}{\partial t}\right|+\ln\left|\frac{\partial t}{\partial y}\right|\nonumber \\
 & =\ln\frac{1}{t(1-t)}+\ln\left|\sum_{i=1}^{k}w_{i}\mathcal{N}(y,\mu_{i},\sigma_{i})\right|\label{eq:ln_det}
\end{align}

\begin{algorithm}[t]
\caption{Causal ordering identification.}
\label{alg:multivariate-ordering}
\KwIn{Data matrix $X\in\mathbb{R}^{n\times d}$.} 
\KwOut{The causal ordering $ \pi $.}

$\pi \gets []$\; 
$\mathrm{nodes} \gets \{1,2,..., d\}$\; 
\Repeat{$\mathrm{nodes} = \emptyset $}
{  \For{$i \in \mathrm{nodes}$}{         
$X_{(-i)} = \{x_j\}_{j \in \mathrm{nodes} \setminus \{i\}} $\;         
Estimate the noise by CAF-PoNo model:          
$$\hat{\epsilon_i} = g^{-1}_{\theta}(x_i) - h_\theta(X_{(-i)})$$\\
       Compute the sink score:         
 	$$S(x_{i}):=-\left(\max_{x\in X_{(-i)}}\mathrm{HSIC}(\hat{\epsilon}_{i},x)\right)$$       }    
$\mathrm{sink} \gets \underset{x_{i}\in\mathrm{nodes}}{\arg\max}S(x_{i})$\;     
 $\pi \gets [\pi, \mathrm{sink}]$\;    
 $\mathrm{nodes}\gets \mathrm{nodes}-\mathrm{\{sink\}}$\;} \KwRet{$\pi$}
\end{algorithm}

By substituting Eq.~\eqref{eq:sigmoid_CDF} and Eq.~\eqref{eq:ln_det} into Eq.~\eqref{eq:mle_loss}, we get a fully differentiable loss function
that can be optimized using off-the-shelf gradient-based methods. 
Moreover, the adoption of the MLE framework in cause-effect estimation is justified by established theory \cite{zhang2015estimation} in the following Lemma.


\begin{lemma} \label{lem:mle} (Maximum likelihood for causal discovery under PNL models, Theorem 2 of \cite{zhang2015estimation}). 
The parameter set $\theta^*$
that maximizes the likelihood $\mathbb{E}[\ln p_\theta(Y\mid X)]$ also minimizes the mutual information of the cause $X$ and the noise $\epsilon_Y$.
\end{lemma}


The proof can be found in \cite{zhang2015estimation}. Simply put, Lemma \ref{lem:mle} ensures that with a sufficiently capable model, maximizing MLE is equivalent to reaching the ground truth PNL in the true direction, given access to the ground truth joint distribution. 

\subsection{Cause-Effect Inference}
When the PNL model is identifiable, by definition, there exists no reverse model $x=g_X(h_X(y)+\epsilon_X)$ such that $g_X$ is invertible and $Y\indep \epsilon_X$ \cite{zhang2009onthe}, which implies that the best model found by MLE in the anti-causal direction will have $\epsilon_X \dep Y$, since invertibility is already guaranteed. This insight enables us to design an independence-based cause-effect identification method.

More specifically, to decide the causal direction, we fit a
model corresponding to each direction $X\rightarrow Y$ and $Y\rightarrow X$ to the data, after which we obtain the estimated noises $\hat{\epsilon}_{Y}$ and $\hat{\epsilon}_{X}$,
respectively. Subsequently,
the model that exhibits independence between the estimated noise and the putative cause is chosen as the more likely causal explanation.
To quantify independence, we employ the Hilbert-Schmidt
independence criterion (HSIC) \cite{gretton2005measuring} which is
a popular kernel-based dependence measure that requires no parameter estimation.
A higher HSIC value indicates a stronger dependence, while a lower
value indicates a lower level of dependence. Consequently, the independence
score is the negative value of the HSIC. The details of the proposed
bivariate causal discovery algorithm is illustrated in Algorithm~\ref{alg:bivariate-causalinference}. 

\subsection{Extension to Multivariate Causal Discovery}

An important aspect of our approach is that it is readily extendable
to multiple variables, thus effectively addressing the challenges
of the multivariate causal discovery problem. 

Towards this end, we decompose the causal structure learning task into two stages, where we first identify the causal ordering among the variables, then the edges are recovered with respect to said ordering. This approach ensures the acyclicity of the resultant graph, while attaining an efficient polynomial runtime.

\subsubsection{Causal Ordering Identification}
\begin{definition}
A causal ordering of a graph $\mathcal{G}$ is a non-unique permutation~$\pi$
of $d$~nodes such that a given node always precedes its descendants
in $\pi$, i.e., $\mbox{\ensuremath{i\prec_{\pi}j\Leftrightarrow j\in\text{Des}_{\mathcal{G}}(i)}}$. 
\end{definition}
Each permutation~$\pi$ corresponds to a unique, fully connected
DAG $\mathcal{G}^{\pi}$ where every pair of nodes $i\prec_{\pi}j$
defines directed edges $(i\rightarrow j)$ in $\mathcal{G}^{\pi}$.
Therefore, the graph $\mathcal{G}$ that we seek for is a subgraph
of $\mathcal{G}^{\pi}$ if $\pi$ is among the topological sorts of $\mathcal{G}$. As a
consequence, a pruning procedure needs to be performed to eliminate
redundant edges from $\mathcal{G}^{\pi}$.

Following \cite{uemura2022amultivariate}, we utilize the two following propositions to find one of causal orders of the underlying causal structure, by recursively detecting and excluding the sink node. 
\begin{proposition}
\label{proposition-ind-noise-1}The noise $\epsilon_{i}$ of every
node $x_{i}$ is independent of all its non-descendants.
\end{proposition}
\begin{proposition}
\label{proposition-sink-node-1} An arbitrary node $x_{i}$ is considered
as a sink node if and only if its corresponding noise $\epsilon_{i}$
is independent of all other variables, which are denoted as $X_{(-i)}:=X\backslash\{x_{i}\}$.
\end{proposition}

We refer to \cite{uemura2022amultivariate} for the detailed proof of these propositions. To identify the sink node, we extend
the proposed bivariate model to accommodate multiple causes. Specifically, the exogenous noise of an arbitrary node $x_i$, denoted as $\epsilon_{i}$, is estimated from the other variables excluding $x_i$, i.e., $\mbox{\ensuremath{\hat{\epsilon}_{i}:=g_{\theta}^{-1}(x_{i})-h_{\theta}(X_{(-i)})}}$.
To check whether $x_{i}$ is a sink node, we introduce the sink score,
which quantifies the independence degree of the estimated noise $\hat{\epsilon}_{i}$
and $X_{(-i)}$. The sink score of a node $x_{i}$,
denoted as $S(x_{i}),$ is defined as the negative of the maximum
value of HSIC between $\hat{\epsilon}_{i}$ and every $x\in X_{(-i)}$:
\begin{equation}
S(x_{i}):=-\left(\max_{x\in X_{(-i)}}\mathrm{HSIC}(\hat{\epsilon}_{i},x)\right)
\end{equation}
By evaluating the sink score for each variable, we can identify the
sink node as the variable with the highest sink score. Once the sink
node is identified, it is removed from the list and the process is
repeated with the remaining variables until the full causal ordering
is identified. The details of the causal ordering process is outlined in Algorithm \ref{alg:multivariate-ordering}.

\subsubsection{Pruning Method \label{subsec:Pruning-Method-1}}

To eliminate spurious edges and refine the causal structure, we employ
an array of conditional independence tests based on a given permutation.
In particular, we remove edges $(i\rightarrow j)$ if $X_{i}\indep X_{j}|X_{\mathrm{pre_{\pi}}(j)\backslash\{i\}}$
where $\mathrm{pre}_{\pi}(j)$ is a set of preceding variables in
the causal ordering $\pi$ \cite{verma1990causalnetworks}. This pruning
method does not depend on the SEM assumption, and yet is more flexible
than the CAM pruning, a popular pruning method in other ordering-based
studies \cite{rolland2022scorematching,sanchez2023diffusion,yang2023reinforcement}, which is however only designed for generalized additive models. That being said, we also investigate both approaches comparatively in the Appendix \cite{hoang2024enabling}.

To perform the conditional independence tests, we employ a recent
state-of-the-art latent representation learning based conditional
independence testing (LCIT) method \cite{duong2022conditional}, which does not make any parametric assumption and is shown to scale linearly with the sample size. It
is worth mentioning that alternative conditional independence testing
methods can also be employed during this pruning process. We summarize the pruning method in Algorithm \ref{alg:multi-prunning}. Together, Algorithm~\ref{alg:multivariate-ordering} and Algorithm~\ref{alg:multi-prunning} highlights the key steps for the CAF-PoNo method in multivariate causal discovery.

\subsubsection{Complexity Analysis }

The causal ordering identification in Algorithm~\ref{alg:multivariate-ordering}
requires $\mathcal{O}(d^2)$ iterations, each spending $\mathcal{O}(nd)$
for the noise estimation using CAF-PoNo and $\mathcal{O}(n^{2}d)$
for the sink score calculation using HSIC, leading to a total
complexity of $\mathcal{O}(n^{2}d^{3})$. Similarly, the pruning step
in Algorithm~\ref{alg:multi-prunning} performs $\mathcal{O}(d^{2})$ conditional
independence tests, each costing $\mathcal{O}(nd)$, resulting to
the final complexity of $\mathcal{O}(nd^{3})$.

\begin{algorithm}[t]
\caption{Pruning algorithm.}
\label{alg:multi-prunning}

\SetNoFillComment
\KwIn{Data matrix $X\in\mathbb{R}^{n\times d}$,  a causal ordering $\pi$, a conditional independence testing method $\mathrm{CIT}$ returning the $p$-value for the hypothesis $X_i \indep X_j | X_Z$, and a significance level $\alpha$ for the test.} 
\KwOut{The adjacency matrix $ A $ of the final DAG.}

Initialize $A$ as a zero value matrix of size $d\times d$.\;
\For{$j \in \pi$}{     
$\mathrm{Pre}_j = \pi[:j]$\;     
\For{$i \in \mathrm{Pre}_j$}{         
$Z = \mathrm{Pre}_j \setminus \{i\}$\;
$A[i,j]=\begin{cases}1 & \text{if }\mathrm{CIT}(X_{i},X_{j},X_{Z})<\alpha\\0 & \text{otherwise}\end{cases}$ \;     
}     
} 
\KwRet{$A$}
\end{algorithm}

%% file: experiments.tex
\begin{table}[t]
\centering
\caption{Bivariate causal discovery performance on various synthetic datasets. The evaluation metric
is AUC (higher is better). The proposed CAF-PoNo method is compared
against ANM \cite{hoyer2008nonlinear}, PNL \cite{zhang2009onthe}, CAREFL \cite{khemakhem2021causalautoregressive}, AbPNL \cite{uemura2020estimation}, RECI \cite{blobaum2018cause}, and LOCI \cite{immer2023onthe}. We highlight in bold the \textbf{best} result and in underline the
\underline{second best} result. CAF-PoNo achieves the highest AUC on most benchmark datasets.\newline}
\resizebox{\columnwidth}{!}{
\begin{tabular}{@{}llllllll@{}}
\toprule
\multicolumn{1}{c}{\multirow{2}{*}{Method}} & \multicolumn{3}{c}{Simple PNL (different noises)}                   & \multicolumn{1}{c}{\multirow{2}{*}{GP PNL}} & \multicolumn{1}{c}{\multirow{2}{*}{LS}} & \multicolumn{1}{c}{\multirow{2}{*}{AN}} & \multicolumn{1}{c}{\multirow{2}{*}{Overall}} \\ \cmidrule(lr){2-4}
\multicolumn{1}{c}{}                       & Gaussian& Laplace& Uniform& \multicolumn{1}{c}{}                        & \multicolumn{1}{c}{}                    & \multicolumn{1}{c}{}                    & \multicolumn{1}{c}{}                         \\ \midrule
ANM                                        & 56.10          & 60.26          & 54.97          & 82.71                                       & 48.78                                   & 53.84                                   & 59.44                                        \\
PNL                                        & {\underline{91.41}}    & 68.42          & 88.64          & 85.71                                       & 67.83                                   & 66.59                                   & 78.10                                        \\
CAREFL                                     & 85.43          & 79.24          & {\underline{89.61}}    & {\underline{90.10}}                                 & 75.53                                   & 82.46                                   & {\underline{83.73}}                                  \\
AbPNL                                      & 45.43          & 51.81          & 47.94          & 13.49                                       & 40.86                                   & 65.66                                   & 44.19                                        \\
RECI                                       & 81.38          & {\underline{89.72}}    & 39.70          & 87.72                                       & 84.91                                   & 91.88                                   & 79.22                                        \\
LOCI                                       & 00.00          & 00.00          & 00.00          & 00.00                                       & \textbf{97.04}                          & \textbf{100.00}                         & 32.84                                        \\
\midrule
Ours                                       & \textbf{95.44} & \textbf{93.51} & \textbf{98.73} & \textbf{95.88}                              & {\underline{89.16}}                             & {\underline{97.17}}                             & \textbf{94.98}                               \\ \bottomrule
\end{tabular}
}
\label{tab:synthetic-results}
\end{table}

\begin{table}[t]
\caption{Bivariate causal discovery performance on real datasets (Tübingen dataset \cite{mooij2016distinguishing}). The evaluation metrics
are AUC  and Accuracy (higher is better). The proposed CAF-PoNo method is compared
against ANM \cite{hoyer2008nonlinear}, PNL \cite{zhang2009onthe}, CAREFL \cite{khemakhem2021causalautoregressive}, AbPNL \cite{uemura2020estimation}, RECI \cite{blobaum2018cause}, and LOCI \cite{immer2023onthe}. We highlight in bold the \textbf{best} result and in underline the
\underline{second best} result. CAF-PoNo attains the highest scores on both AUC and accuracy.
}
\centering{}%
\resizebox{\columnwidth}{!}{
\begin{tabular}{cccccccc}
\toprule 
Method & ANM & PNL & AbPNL & CAREFL  & RECI & LOCI & \textbf{CAF-PoNo} \tabularnewline
\midrule 
AUC & 58.86\% & 63.22\% & 68.51\% & 65.10\% & \underline{69.67\%} & 62.04\% & \textbf{73.84\%}\tabularnewline
\midrule 
Accuracy & 52.52\%	& 59.59\% &	62.62\% &	52.52\% &	62.62\% & 	\underline{63.63\%} &	\textbf{71.71\%} \tabularnewline
\bottomrule
\end{tabular}
}
\label{tab:Results-on-Tuebingen}
\end{table}


\section{Numerical Evaluations} \label{sec:Experiments}

To assess the quality of the proposed method, we compare it with state-of-the-art
causal discovery models on two scenarios: bivariate and multivariate
cause-effect inference, using both synthetic and real data, where the ground truth causal directions/structures are available.

Regarding parameter selection in our method, for each dataset, the hyperparameters are chosen to maximize the AUC score, with values among the following sets:
\begin{itemize}
\item The number of Gaussian components $k$: $\{4,8,10,12\}$.
\item The hidden size of a single layer neural network: $\{16,32,64,128\}$.
\item The number of epochs: $\{200,300,500,700,1000\}$.
\end{itemize}

All parameters are trained using the Adam optimizer with early stopping,
learning rate of 0.001, and batch size of 128. We randomly split
the datasets into training and test sets with a ratio of 6:4. The
validation set comes from 20\% of the training set. The training set
is used to learn model parameters, while the validation set is utilized
for early stopping. The testing set is for computing the independence
score for causal direction identification. Following previous studies \cite{zhang2009onthe,uemura2020estimation,uemura2022amultivariate},
we use the same setting for HSIC, including
Gaussian kernel and bandwidth $\sigma$ of 1. 

\subsection{Bivariate Causal Discovery}

\subsubsection{Baselines and the Evaluation Metric}

We compare the bivariate causal discovery performance of the proposed method against several popular baselines as
described below: 
\begin{itemize}
\item Additive noise model (ANM) \cite{hoyer2008nonlinear} estimates the
causal mechanism via Gaussian Process regressions and utilizes HSIC
to assess the independence between the cause and residuals. 
\item Post nonlinear (PNL) \cite{zhang2009onthe} reconstructs the noise
using two neural networks trained to minimize the mutual information
between the putative cause and the noise, and utilizes HSIC test for
independence test between the cause and the estimated noise. 
\item Causal Autoregressive Flow (CAREFL) \cite{khemakhem2021causalautoregressive}
utilizes affine normalizing flows to estimate the affine causal
model introduced in the paper. 
\item Autoencoder-based post nonlinear (AbPNL) \cite{uemura2020estimation}
models PNL using an auto-encoder trained with a combination of independence
and reconstruction losses. 
\item RECI \cite{blobaum2018cause} employs a monomial regressor to predict the target variable using normalized versions of potential causal variables and then compares the mean squared errors (MSE) of these regressions for each possible causal direction. The direction with the lower MSE is selected as the inferred causal relationship. 
\item LOCI \cite{immer2023onthe} harnesses the estimation of location scale noise models (LSNM) via heteroscedastic regression to reconstruct the underlying noise in the observed data. The causal direction is then determined by comparing the level of independence between the estimated noise and the presumed cause under two possible causal directions.
\end{itemize}

We utilize code from Causal Discovery Toolbox (CDT)\footnote{\url{https://github.com/FenTechSolutions/CausalDiscoveryToolbox}} and Causal-learn framework\footnote{\url{https://github.com/py-why/causal-learn}} for ANM, RECI and PNL methods. The implementations of the other methods are derived from the original code published by the authors.

As for the evaluation metric, we utilize the bi-directional area under
the receiver operating characteristic curve (AUC), where higher values
indicate better performance. The AUC was computed by averaging two
scores: one measures classification performance for determining if
a pair $(X,Y)$ belongs to the $X\rightarrow Y$ class, and the other
accounts for the $Y\rightarrow X$ class.

\subsubsection{Synthetic data}
We generate datasets that follow the PNL model in Eq. \eqref{eq:PNL}, considering both simple and complex functional relationships. Following \cite{zhang2009onthe, uemura2020estimation}, simple functional relationships are captured in the Simple-PNL dataset, which is constructed using various function choices as outlined in Table~\ref{tab:synthetic-data} of the Appendix \cite{hoang2024enabling}. We consider three different noise distributions: Gaussian,
Uniform, and Laplace noises. For each distribution,
we generate 500 pairs of $(X,Y)$, each pair consisting of 1,000 i.i.d. samples. Furthermore, we introduce PNL-GP dataset with more intricate functional relationships within the PNL model where $h\left(.\right)$ and $g\left(.\right)$ are the weighted sum of Gaussian Processes and sigmoid functions, respectively. To assess CAF-PoNo's effectiveness beyond the PNL assumption, we also utilize synthetic datasets generated from non-linear additive noise (AN) and location-scale (LS) noise models, as provided by \cite{tagasovska2020distinguishing}.


The results are presented in Table~\ref{tab:synthetic-results}, demonstrating that 
CAF-PoNo outperforms other methods on all PNL datasets, achieving the highest AUC score overall. Notably, the results highlight the robustness of CAF-PoNo across various types of noise even though the base noise type is solely set as Gaussian, implying CAF-PoNo's effectiveness in addressing noise mis-specification. Interestingly, the results also reveal CAF-PoNo's capability in handling model mis-specification. In more details, CAF-PoNo attains an AUC of approximately 90\% under LS data, demonstrating a comparable performance to LOCI, a method specifically designed for the LS scenario.


\subsubsection{Real world data}

To demonstrate the effectiveness of CAF-PoNo in real world scenarios, we utilize
the Tübingen dataset \cite{mooij2016distinguishing}, a common benchmark in bivariate causal discovery comprising of 99 cause-effect pairs from a variety
of domains, e.g., abalone measurements, census income, and liver disorders.
The number of samples for each dataset varies from 94 to 16,382.

Table~\ref{tab:Results-on-Tuebingen} illustrates the comparative
performance in terms of AUC and Accuracy scores. The results clearly demonstrate that
CAF-PoNo outperforms other baselines on both metrics when applied to real-world scenarios.
Specifically, CAF-PoNo achieves the highest AUC, surpassing the runner-up RECI model by a significant margin of around 5\%. Following prior studies \cite{khemakhem2021causalautoregressive, immer2023onthe}, we also provide the accuracy which is a common metric in bivariate causal discovery. As evidenced by the table, CAF-PoNo again accomplishes the highest accuracy of 71.71\%, leaving the second-best method LOCI by a large margin of over 8\%.  


\subsection{Multivariate Causal Discovery}

\begin{figure*}[t]
\includegraphics[width=1\textwidth]{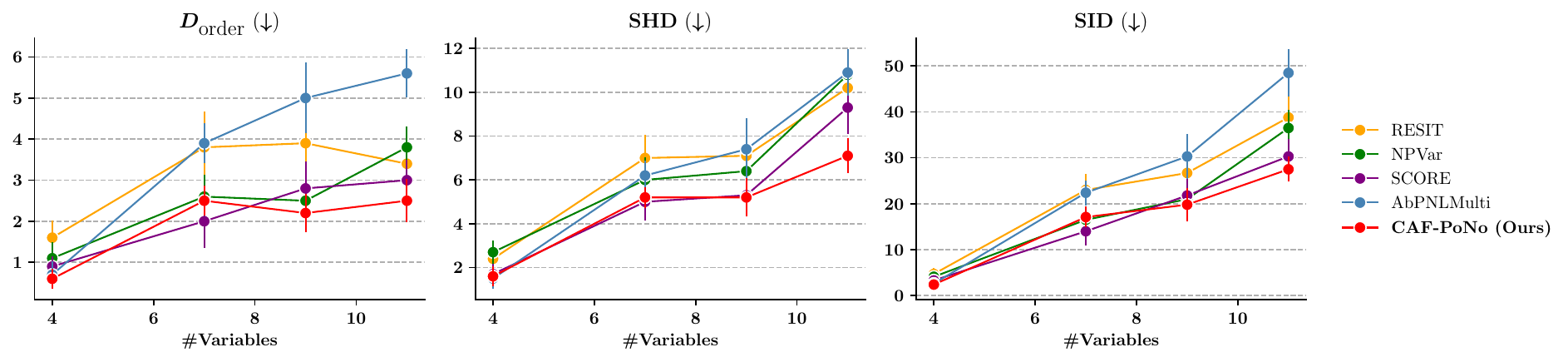}\caption{Multivariate causal discovery performance on synthetic data as a function
of the number of variables. We fix $n=1,000$ and vary the number
of variables. The evaluation metrics are $D_{\mathrm{order}}$, SHD,
and SID (lower is better). The reported values are aggregated over
10 independent runs. We compare the proposed CAF-PoNo method with
RESIT \cite{peters2014causaldiscovery}, NPVar \cite{gao2020apolynomialtime},
SCORE \cite{rolland2022scorematching}, and AbPNLMulti \cite{uemura2022amultivariate}.\newline}
\label{fig:multi-numvars}
\end{figure*}

\begin{figure*}[t]
\includegraphics[width=1\textwidth]{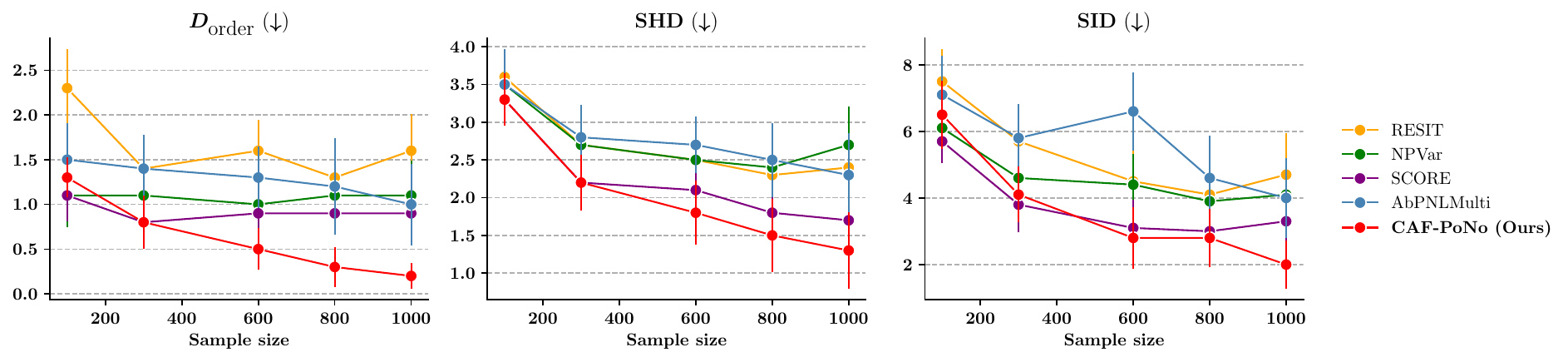}\caption{Multivariate causal discovery performance on synthetic data as a function
of sample size. We fix $d=4$ and vary the sample size. The evaluation
metrics are $D_{\mathrm{order}}$, SHD, and SID (lower is better).
The reported values are aggregated over 10 independent runs. We compare
the proposed CAF-PoNo method with RESIT \cite{peters2014causaldiscovery},
NPVar \cite{gao2020apolynomialtime}, SCORE \cite{rolland2022scorematching},
and AbPNLMulti \cite{uemura2022amultivariate}.\newline}
\label{fig:multi-samplesize}
\end{figure*}

\subsubsection{Baselines and Evaluation Metrics}

We consider the multivariate version of AbPNL \cite{uemura2022amultivariate}
which is the first work demonstrating multivariate causal discovery
for PNL model. In addition, we also include three ordering-based
methods designed for ANM models including RESIT \cite{peters2014causaldiscovery},
NPVar \cite{gao2020apolynomialtime} and SCORE \cite{rolland2022scorematching}.
The causal orders obtained from these methods are applied with the
same pruning procedure as described in Section \ref{subsec:Pruning-Method-1}
with a commonly adopted significance level of 0.001. 

We consider two common structural
metrics for multivariate causal discovery evaluation: the structural Hamming distance (SHD) and structural intervention
distance (SID) \cite{peters2015structural}. These metrics enable
us to quantify the error between the predicted DAG with the true DAG. In particular, the
SHD measures the total number of missing, extra, and reverse
edges between the estimate and the true DAG, while the SID measures
the minimum number of interventions required to transform the output
to the true DAG. Hence, lower values for both SHD and SID indicate
a better fit of the output DAG to the given data. In addition, we
also consider the order divergence $(D_{\mathrm{order}})$ \cite{rolland2022scorematching}
as a measure of the quality of the causal ordering. More specifically,
the order divergence reflects the number of directed edges in the
true DAG that disagree with the causal ordering.

\subsubsection{Synthetic data}

We generate Erdös-Rényi
causal graphs \cite{bollobas1984theevolution} with $d$ nodes with an expected in-degree of 2. The data is generated
following the PNL model \eqref{eq:PNL_multi-1}, where $h_{i}$ represents
the weighted sums of Gaussian processes and $g_{i}$ represents sigmoid
functions. For each causal structure, we generate 1,000 samples from
uniform noises $\mathcal{U}(0,1)$. 

\textbf{Effect of dimensionality. }To study the performance of
the proposed method across different numbers of nodes, we vary $d$ from 4 to 11 variables. The results are shown in Figure~\ref{fig:multi-numvars}, showing that our proposed method consistently outperforms
other baselines in most cases. Remarkably, the proposed method shows
a significantly low $D_{\mathrm{order}}$ compared with competitors when the graph size grows.
Surprisingly, AbPNLMulti exhibits the poorest performance although
the method is specifically designed for the PNL model.

\textbf{Effect of sample size. }In order to assess the impact of sample
sizes on the performance of different models, we vary the sample size from 100 to 1,000 while fixing $d=4$. The results in Figure~\ref{fig:multi-samplesize}
reveal that CAF-PoNo constantly surpasses other methods across various
settings. In these cases, CAF-PoNo achieves extraordinarily low $D_{\mathrm{order}}$
values, hovering around 1. Notably, CAF-PoNo exhibits exceptional
efficiency when the sample size exceeds 500. In particular, both SHD
and SID decrease to below 2 and 4, respectively. These results underscore
the robustness and effectiveness of CAF-PoNo, particularly its sample efficiency.

\begin{figure}[t]
\centering
\includegraphics[width=\columnwidth]{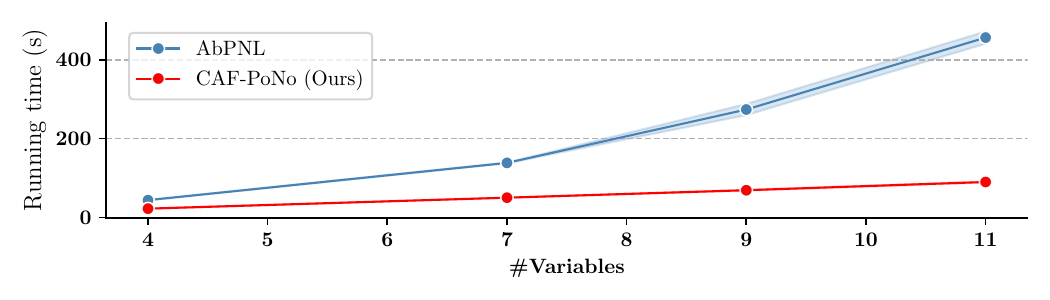}\caption{The running time in seconds as a function of the number of variables. The proposed method exhibits a significant reduction in running time compared to AbPNL, demonstrating its potential scalability to high dimensional data.\newline\vspace{1mm}}
\label{fig:running_time}
\end{figure}

\textbf{Running time.} Figure \ref{fig:running_time} illustrates the running time of both AbPNL and CAF-PoNo algorithms as the number of variables increases. The results clearly demonstrate that CAF-PoNo exhibits a substantial reduction in running time with up to 5x speed-up compared to AbPNL. This considerable gain stems from the lightweight of the CDF flow and our proposed MLE framework, which eliminates the need for mutual information calculations in AbPNL, leading to a more computationally efficient model. 

\subsubsection{Real data}

We test these methods on the Sachs dataset \cite{sachs2005causalproteinsignaling},
a popular benchmark dataset in multivariate causal discovery. The dataset contains 853
observational samples, representing expression data of the protein
signaling network of 11 nodes and 17 edges. The results are shown
in Table~\ref{tab:Results-on-Sachs}, which demonstrates the superior
performance of our method in comparison with state-of-the-art baselines, evidenced by lowest SHD and SID among all methods. These results highlight
the remarkable capability of our proposed method in successfully uncovering
intricate causal structures in real-world datasets.

\begin{table}[t]
\caption{Multivariate causal discovery performance on a real data of the protein
signaling network. The evaluation metrics are SHD, and SID (lower
is better). We compare the proposed CAF-PoNo method with four baselines
RESIT \cite{peters2014causaldiscovery}, NPVar \cite{gao2020apolynomialtime},
SCORE \cite{rolland2022scorematching}, and AbPNLMulti \cite{uemura2022amultivariate}.}
\label{tab:Results-on-Sachs}
\centering{}%
\resizebox{.65\columnwidth}{!}{
\begin{tabular}{ccc}
\toprule 
Method & SHD ($\downarrow$) & SID ($\downarrow$)\tabularnewline
\midrule 
RESIT & 13 & 47\tabularnewline
NPVar & 14 & 57\tabularnewline
SCORE & 12 & 45\tabularnewline
AbPNLMulti & 13 & 47\tabularnewline
\midrule 
\textbf{CAF-PoNo (Ours)} & \textbf{11} & \textbf{41}\tabularnewline
\bottomrule
\end{tabular}
}
\end{table}

%% file: conclusion.tex
\section{Conclusion}
In this paper, we introduce CAF-PoNo, an innovative
and efficient flow-based approach for cause-effect identification
under the PNL model. Moreover, CAF-PoNo surpasses existing methods
in parameter estimation for the PNL model by adhering to the invertibility
constraint, resulting in the highest AUC on synthetic and real datasets.
We also extend CAF-PoNo for multivariate causal discovery and show
its superiority over state-of-the-art methods.

%% file: appendix.tex
\section*{Appendix for ``Enabling Causal Discovery in Post-Nonlinear Models with Normalizing Flows''}

\begin{table}[h]
\caption{Parameter choices for synthetic data generation
include various functional forms of two transformations and noise
distributions for the PNL model.}
\begin{centering}
\emph{}%
\begin{tabular}{c|l}
\hline 
\emph{Parameter} & \emph{Choices}\tabularnewline
\hline 
\multirow{6}{*}{\emph{$g$}} & Linear: $g(x)=a*x$\tabularnewline
\cline{2-2} 
 & Cube: $g(x)=x^{3}$\tabularnewline
\cline{2-2} 
 & Inverse: $g(x)=\frac{1}{x}$\tabularnewline
\cline{2-2} 
 & Exp: $g(x)=\exp(x)$\tabularnewline
\cline{2-2} 
 & Log: $g(x)=\log(x)$\tabularnewline
\cline{2-2} 
& Sigmoid: $g(x)=\frac{1}{1+e^{-x}}$\tabularnewline
\hline
\multirow{3}{*}{\emph{$h$}} & Square: $h(x)=x^{2}$\tabularnewline
\cline{2-2} 
 & Absolute: $h(x)=\left|x\right|$\tabularnewline
 \cline{2-2}
& Sigmoid: $h(x)=\frac{1}{1+e^{-x}}$\tabularnewline
\hline 
\multirow{3}{*}{\emph{$p(\epsilon)$}} & $\mathcal{N}(0,1)$\tabularnewline
\cline{2-2} 
 & $\mathcal{U}(-1,1)$\tabularnewline
\cline{2-2} 
 & $\mathrm{Laplace}(0,1)$\tabularnewline
\hline 
\end{tabular}
\par\end{centering}
\label{tab:synthetic-data}
\end{table}

%
%


\subsection{Additional Experiments}

\begin{figure}[h]
\centering
\includegraphics[width=0.7\textwidth]{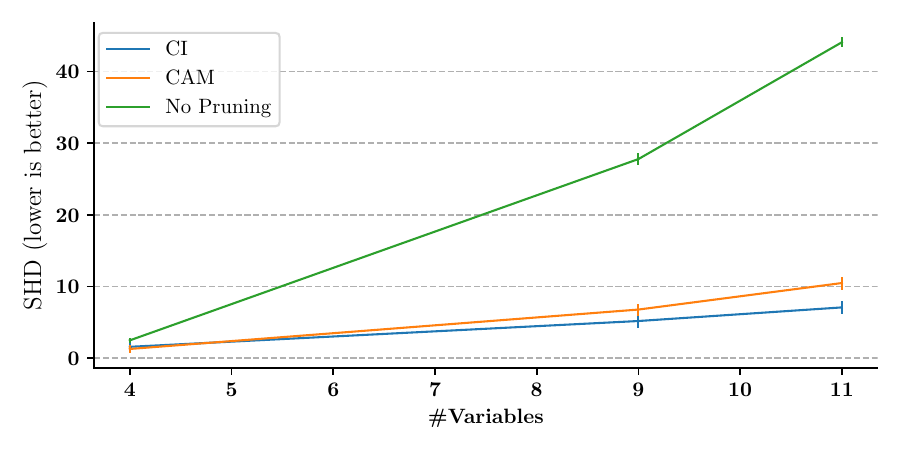}\caption{The performance of the proposed method on multivariate causal structure learning with different pruning approaches as a function of the number of variables in terms of SHD (lower is better). The pruning approaches includes the no-pruning approach, the conditional independence test based approach (CI), and the causal additive model (CAM) approach.\newline}
\label{fig:pruning_methods}
\end{figure}

\noindent \textbf{Pruning methods.} To evaluate the effectiveness of the pruning method, we have compared the SHD scores (lower is better) obtained with and without pruning methods like the CI tests-based approach and another common technique known as the CAM pruning algorithm [9] in Figure \ref{fig:pruning_methods}. The results demonstrate that incorporating pruning methods leads to significantly better SHD scores. This improvement stems from the ability of pruning to eliminate spurious edges that do not reflect true causal relationships. Furthermore, the pruning approach based on CI tests outperforms the CAM algorithm in terms of SHD, highlighting the effectiveness of CI tests in identifying irrelevant edges within the PNL model.

